\pdfoutput=1
\documentclass[twocolumn]{article}
\usepackage{graphicx} 
\usepackage{todonotes}
\usepackage[style=numeric, sorting=none, sortcites=true]{biblatex}
\usepackage{enumitem}
\usepackage[table]{xcolor}
\usepackage{amsmath}
\usepackage{amssymb}
\usepackage{authblk}
\usepackage[margin=1in]{geometry}
\usepackage{booktabs}
\usepackage{multirow} 
\usepackage{microtype}
\usepackage[autostyle]{csquotes}

\newcommand{\etal}{\textit{et al.}}

\newcommand{\infonce}{\mathrm{InfoNCE}}
\newcommand{\rmse}{\mathrm{RMSE}}
\newcommand{\gt}{\mathrm{GT}}
\newcommand{\patch}{\mathrm{patch}}
\newcommand{\CLS}{\mathrm{CLS}}
\newcommand{\emb}{\mathrm{emb}}
\newcommand{\myalign}{\mathrm{align}}
\newcommand{\depth}{\mathrm{depth}}

\newcommand{\figref}[1]{Fig.~\ref{#1}}
\newcommand{\tabref}[1]{Table~\ref{#1}}

\title{\textbf{PureCLIP-Depth: Prompt-Free and Decoder-Free Monocular Depth Estimation within CLIP Embedding Space}}
\author{
    Ryutaro Miya \quad Kazuyoshi Fushinobu \quad Tatsuya Kawaguchi \\
    Institute of Science Tokyo \\
    \small \texttt{miya.r.7e86@m.isct.ac.jp}
}
\date{}

\begin{document}

\maketitle

\begin{center}
    \section*{Abstract}
\end{center}
We propose PureCLIP-Depth, a completely prompt-free, decoder-free Monocular Depth Estimation (MDE) model that operates entirely within the Contrastive Language-Image Pre-training (CLIP) embedding space. Unlike recent models that rely heavily on geometric features, we explore a novel approach to MDE driven by conceptual information, performing computations directly within the conceptual CLIP space. The core of our method lies in learning a direct mapping from the RGB domain to the depth domain strictly inside this embedding space. Our approach achieves state-of-the-art performance among CLIP embedding-based models on both indoor and outdoor datasets.
The code used in this research is available at: \url{https://github.com/ryutaroLF/PureCLIP-Depth}

\section{Introduction}
Monocular Depth Estimation (MDE) aims to reconstruct the three-dimensional structure of a scene from a single RGB image and serves as a fundamental technology in various domains such as autonomous driving~\cite{godard2019digging}, robotics~\cite{zhou2017unsupervised}, augmented and virtual reality (AR/VR)~\cite{li2019learning}, and environmental perception~\cite{cordts2016cityscapes}.
Despite its wide applicability, the task remains inherently ill-posed, since a monocular image does not contain absolute scale information~\cite{hartley2003multiple}.
In the early stage of deep learning–based MDE, Eigen \etal~\cite{eigen2014depth} demonstrated the first direct prediction of depth from a single image using a multi-scale convolutional neural network (CNN), which laid the foundation for subsequent CNN-based approaches~\cite{laina2016deeper, fu2018deep}.
More recently, the introduction of Vision Transformers (ViTs) further advanced MDE, as demonstrated by MiDaS~\cite{ranftl2020towards} and DPT~\cite{ranftl2021vision}.
While transformer-based models have significantly improved the representational power of MDE, the information they exploit remains largely limited to visual and geometric features.
Beyond visual features, human depth perception relies heavily on semantic context. This observation leads to the hypothesis that the vast world knowledge embedded in language models can provide essential priors for the physical scale and spatial relationships of objects.
Consequently, there is a growing trend to utilize language and vision-language models (VLMs) as a rich source of semantic information to equip models with a deeper understanding of the three-dimensional environment.

For example, WorDepth~\cite{zeng2024wordepth} introduced the concept of a variational language prior, which models language as a probabilistic prior distribution over depth for the first time.
Specifically, their approach utilizes image captions to learn the latent distribution of three-dimensional scenes and leverages language and vision in a complementary manner to achieve metric-scale depth estimation.
Subsequent studies~\cite{zhang2025vision, zhang2025language} have extended this language-based prior by incorporating camera models and physical constraints into the learning framework.
In recent years, approaches incorporating more advanced language understanding have been proposed, including methods that utilize large language models (LLMs)~\cite{xia2024large} and those that combine language embeddings from BERT with semantic segmentation~\cite{auty2024language}.
Another significant development is the work by Zhang~\etal~\cite{zhang2022can}, who demonstrated that Contrastive Language-Image Pre-training (CLIP)~\cite{radford2021learning} model can perform zero-shot depth estimation by using human language prompts such as \enquote{This object is near/far}. This study stands as a key milestone that first highlighted the potential of CLIP-based depth estimation without explicit re-training.

A common characteristic of these methods is their reliance on explicit textual descriptions or prompts to bridge the gap between language and depth.
While these studies demonstrate a new direction for incorporating language information into depth estimation, they also suffer from a strong dependence on prompt design.
Recent studies have demonstrated that differences in prompt wording or style significantly affect model outputs, often leading to degraded accuracy~\cite{chatterjee2024robustness}.
This issue represents a common limitation among approaches that explicitly treat language information as textual input.

To address this issue, Auty and Mikolajczyk~\cite{auty2023learning} introduced learnable prompt tokens to represent depth as continuous embedding vectors, moving beyond the constraints of natural human language. Their analysis suggested the existence of internal, non-linguistic depth representations within CLIP that do not necessarily align with human language.
Furthermore, Hu~\etal~\cite{hu2024learning} extended zero-shot estimation to a few-shot setting by introducing scene-dependent learnable depth codebooks and text prompts. This approach achieved high-precision and versatile depth estimation across diverse environments.

Prior studies confirm that specific regions within the CLIP embedding space encode depth information. Building on this, we hypothesize the existence of a mapping function that directly aligns RGB embeddings with these depth-relevant regions. Specifically, RGB vectors from the CLIP encoder are transformed into position vectors that match the depth direction in the same space. Inspired by the mapping approach~\cite{wang2024infrared}, we posit that this spatial rotation can be approximated through learnable nonlinear transformations.

While Kim~\etal~\cite{kim2025clip} also explores CLIP-based depth estimation, it relies on intermediate features, shallow features, from the CLIP encoder. This makes the approach more akin to geometry-dependent MDE rather than a pure utilization of the conceptual embedding space. Furthermore, its reliance on a decoder for depth map reconstruction precludes a direct analysis of the relationships within the CLIP space itself.

In contrast, our study aims to estimate depth using only the final output of the CLIP encoder, relying strictly on the conceptual representations learned by CLIP. Unlike WorDepth~\cite{zeng2024wordepth}, which depends on image-level captions, we adopt a patch-based structure to preserve spatial information, balancing local semantics with spatial consistency. Moreover, we simultaneously learn a mapping function from RGB to depth while identifying positions within the CLIP embedding space that correspond to real-world depth values. This joint optimization strategy achieves both conceptual consistency and physical metric accuracy.

The contributions of this study are summarized as follows:
\begin{itemize}[itemsep=1.5ex, topsep=1ex, parsep=0pt]
    \item We empirically demonstrate the existence of a mapping from the RGB domain to the depth domain within the CLIP embedding space.
    \item We achieve metric-scale MDE by performing computations entirely within the conceptual CLIP embedding space without a decoder. This configuration achieves state-of-the-art performance among CLIP encoder-only architectures.
    \item We propose a joint optimization strategy that simultaneously optimizes the RGB-to-depth mapping and metric-scale alignment within the CLIP embedding space.
\end{itemize}

\section{Method}
\subsection{Input: Patch-wise RGB Image}
As mentioned earlier, previous language-guided depth estimation approaches typically attach a single caption to an entire image, which inevitably causes the loss of spatial constraints and positional information during the language encoding process.
To address this limitation, our initial idea was to divide an RGB image into fine-grained patches and convert each patch into a separate language representation.
However, since the CLIP model based on a ViT inherently processes images as a set of patches,
we found it more efficient to directly utilize this built-in patch representation.
In the CLIP encoder, each patch is independently encoded into an embedding vector, which is later aggregated just before the final layer.
By extracting these pre-aggregation patch embeddings, we effectively obtain patch vectors that are already aligned within the CLIP embedding space.
The patch size was set to $14 \times 14$, consistent with the default configuration of the pre-trained ViT-L/14@336px model.
Although smaller patch sizes could provide finer spatial resolution, this would require retraining the entire encoder, which falls outside the scope of this study.
Moreover, overly small patches tend to lose semantic meaning within the CLIP embedding space, as they no longer correspond to recognizable visual concepts.

\subsection{Output: Depth Map}
The proposed model is trained to predict metric-scale depth, which is estimated for each patch individually.
For supervision, the ground-truth depth maps $D_{\gt}$ provided by each dataset are used.
Specifically, for each RGB patch $i$, we compute a representative target depth value $d_i^{\gt}$ by averaging the depth values of all valid pixels within its corresponding spatial region $\Omega_i$. This process is formulated as:
\begin{equation}
d_i^{\gt} = \frac{1}{|\mathcal{V}_i|} \sum_{p \in \mathcal{V}_i} D_{\gt}(p) \, ,
\end{equation}
where $\mathcal{V}_i$ denotes the set of valid pixels within the patch region that satisfy the dataset-specific constraints:
\begin{equation}
\begin{split}
\mathcal{V}_i = \{ p \in \Omega_i \mid \ & d_{\min} \le D_{\gt}(p) \le d_{\max}, \\
& D_{\gt}(p) \neq \text{NaN} \} \, .
\end{split}
\end{equation}
In this formulation, $D_{\gt}(p)$ represents the ground-truth depth value at pixel $p$, while $d_{\min}$ and $d_{\max}$ denote the valid depth range for the respective dataset. The set $\mathcal{V}_i$ filters out invalid or missing values, and $|\mathcal{V}_i|$ represents the count of such valid pixels. The specific values for these thresholds and the detailed masking strategy used during training are described in Section~\ref{sec:datasets}.

To predict depth values corresponding to these targets, the model estimates the metric depth $\hat{d}_i$ by calculating the similarity between the rotated patch embedding $\tilde{\mathbf{z}}_i$ and the entries of a learnable depth table $\mathcal{W}$.
Specifically, the similarity score $s_{ij}$ is defined as:
\begin{equation}
s_{ij} = \frac{\tilde{\mathbf{z}}_i^\mathsf{T} \mathbf{w}_j}{\tau}
\end{equation}
where $\tau$ is a temperature parameter controlling the sharpness of the similarity distribution.
These scores are then normalized by a softmax function to obtain the probability that the $i$-th patch belongs to each depth bin:
\begin{equation}
p_{ij} = \frac{\exp(s_{ij})}{\sum_k \exp(s_{ik})} \, .
\end{equation}

Finally, the expected depth value $\hat{d}_i$ for each patch is estimated as the weighted average of the bin center values $c_j$ according to the probabilities $p_{ij}$:
\begin{equation}
\hat{d}_i = \sum_{j=1}^{K} p_{ij}  c_j \, .
\end{equation}
This softmax-based formulation enables the model to represent depth in a continuous manner, allowing smoother and more fine-grained depth estimation beyond discrete bin boundaries.

\subsection{Learnable Depth Table}
Inspired by Auty and Mikolajczyk~\cite{auty2023learning}, we represent depth bins as learnable embedding vectors rather than fixed values or static language embeddings.
The depth table $\mathcal{W}$ is a set of $K$ depth bins, where each bin is represented as a learnable embedding vector:
\begin{equation}
\mathcal{W} = \{ \mathbf{w}_1, \mathbf{w}_2, \dots, \mathbf{w}_K \} \, .
\end{equation}
Each vector $\mathbf{w}_k \,(k=1, \dots, K)$ was initialized using CLIP text embeddings of phrases such as
“1 meter”, “2 meters”, \dots, “10 meters”.
This initialization embeds the ordinal relationship of depth values within a semantically meaningful region of the CLIP text space, which random initialization cannot capture.

\subsection{Rotating Vectors in CLIP Embedding Space}
Given an input image $I$, the feature vectors $\mathbf{z}$ are extracted from the final layer of the CLIP visual encoder $f(\cdot)$ as:
\begin{equation}
    \mathbf{z} = f(I) \in \mathbb{R}^{B \times (N_{\patch}+1) \times D} \, ,
\end{equation}
where $B$ denotes the batch size, $N_{\patch}$ is the number of patches, $+1$ corresponds to the CLS token, and $D$ is the embedding dimension.
Focusing on one of the patch vectors,
\begin{equation}
\mathbf{z}_{\patch,i} \in \mathbb{R}^D \quad (i = 1, \dots, N_{\patch}) \, ,
\end{equation}
this vector $\mathbf{z}_{\patch,i}$ is then projected into the depth representation space using the mapping function $\phi$:

\begin{equation}
\mathbf{z}'_{i} = \phi\left(\mathbf{z}_{\patch,i}\right), \quad \phi : \mathbb{R}^D \to \mathbb{R}^D \, .
\end{equation}
Here, $\phi$ is implemented as a simple two-layer Multi-Layer Perceptron (MLP), which applies a learnable transformation to rotate the CLIP embedding toward the depth-relevant subspace.

While $\phi$ transforms local features into a depth-relevant space, local patches often remain ambiguous when viewed in isolation. 
This is because the perceived depth of a feature fundamentally depends on the overall scene context. 
For example, a visual feature identified as a ``stone'' is likely a small pebble when located on the nearby ground, whereas the same feature on a distant mountain slope may represent a massive boulder. 

To resolve these ambiguous cases, we incorporate global contextual information by leveraging the CLS token $\mathbf{z}_{\CLS} \in \mathbb{R}^D$ extracted from the same layer as the patch embeddings. 
We concatenate the projected patch embedding $\mathbf{z}'_i$ with $\mathbf{z}_{\CLS}$ and pass them through a fusion function $\psi$ to obtain the final fused representation $\tilde{\mathbf{z}}_i$:
\begin{equation}
    \tilde{\mathbf{z}}_i = \psi\left([\mathbf{z}'_i ; \mathbf{z}_{\CLS}]\right), \quad \psi : \mathbb{R}^{2D} \to \mathbb{R}^D \, .
\end{equation}
Similarly to $\phi$, the function $\psi$ is implemented as a two-layer MLP that learns to integrate local semantics with global scene understanding.

\subsection{Alternating Optimization}
In this study, we adopted a two-phase alternating optimization strategy, consisting of the \textit{Embedding phase} and the \textit{Depth phase}. First, in the Embedding phase, the model learns representations within the embedding space by minimizing the following loss, where the parameters $\phi$, $\psi$, and the depth table $\mathcal{W}$ are updated:
\begin{equation} 
\mathcal{L}_{\emb} = \mathcal{L}_{\myalign}(\tilde{\mathbf{z}}_i, \mathcal{W}) + \lambda_{\infonce} \mathcal{L}_{\infonce}(\tilde{\mathbf{z}}_i, \mathcal{W}) \, .
\end{equation}

Next, in the Depth phase, the model focuses on improving regression accuracy in the metric depth space by minimizing the following loss:
\begin{equation}
\mathcal{L}_{\depth}
=  \mathcal{L}_{\rmse}(\hat{D}, D_{\gt}) \, .
\end{equation}
During this phase, the depth table $\mathcal{W}$ is frozen and used only for reference, while only $\phi$ and $\psi$ are optimized.
This configuration allows the optimization to focus on refining the mapping within the CLIP embedding space itself.

\subsection{Loss Functions}
To optimize each phase effectively, we employed three loss functions: InfoNCE, alignment, and RMSE losses.
The following describes the formulation of each component.
\subsubsection{InfoNCE Loss}
\begin{equation}
\begin{split}
\mathcal{L}_{\mathrm{\infonce}} = & - \frac{1}{\sum_{i} m_i} \sum_{i=1}^{N_{\patch}} m_i \\
& \cdot \log \frac{\exp((\tilde{\mathbf{z}}_i \cdot \mathbf{w}_{y_i}) / \tau)}{\sum_{j=1}^{K} \exp((\tilde{\mathbf{z}}_i \cdot \mathbf{w}_j) / \tau)} \, ,
\end{split}
\end{equation}
where $y_i$ is the ground-truth depth bin index of patch $i$,
$m_i$ denotes the binary mask indicating valid patches, and
$\tau$ is a temperature parameter controlling the sharpness of the similarity distribution.
The InfoNCE loss encourages the fused representation $\tilde{\mathbf{z}}_i$ to be discriminatively aligned with its corresponding depth table vector $\mathbf{w}_{y_i}$ while being separated from other non-corresponding vectors in $\mathcal{W}$.
\subsubsection{Alignment Loss}
The alignment loss is defined as:
\begin{equation}
    \mathcal{L}_{\myalign} = \frac{1}{\sum_{i} m_i} \sum_{i=1}^{N_{\patch}} m_i \left[ 1 - \tilde{\mathbf{z}}_i^\mathsf{T} \mathbf{w}_{y_i} \right] \, ,
\end{equation}
where $m_i$ is a binary mask indicating valid patches. Unlike the InfoNCE loss, which contrasts positive pairs against negative samples, the alignment loss directly maximizes the cosine similarity between each fused representation $\tilde{\mathbf{z}}_i$ and its corresponding target depth vector $\mathbf{w}_{y_i}$. This enforces a strict one-to-one correspondence between the RGB and depth representations in the CLIP embedding space.
\subsubsection{RMSE Loss}
The RMSE loss is computed between the reconstructed depth map $\hat{D}$ and the ground-truth depth map $D_{\gt}$ on a pixel-wise basis:
\begin{equation}
    \mathcal{L}_{\rmse} = \sqrt{ \frac{1}{|\mathcal{V}|} \sum_{p\in\mathcal{V}} (\hat{D}_p - D_{\gt,p})^2 } \, .
\end{equation}
This loss trains the mapping function to accurately project RGB embeddings into the metric depth space while penalizing large prediction errors.

\section{Experiment}
\subsection{Datasets}
\label{sec:datasets}
In this study, we evaluated the performance of the proposed method on both an indoor dataset, NYU Depth V2~\cite{silberman2012indoor}, and an outdoor dataset, KITTI~\cite{geiger2012we}.
For NYU Depth V2, the training and testing sets followed the official splits provided by Lee~\etal~\cite{lee2019big}. The training set was further divided into training and validation subsets with a 9:1 ratio through random sampling. This split was performed only once and kept fixed throughout all experiments.
All RGB images were resized to $336 \times 336$ using bilinear interpolation. Following the Eigen crop setting~\cite{eigen2014depth}, invalid border regions in the ground-truth depth maps were excluded from both evaluation and loss computation. Additionally, during training, pixels within a patch were excluded from the loss calculation if more than half of them had missing depth values. 
Specifically for the NYU Depth V2 dataset, the depth table was initialized with 15 learnable vectors corresponding to 15 equally spaced depth bins covering the range from 0 to 10 meters. These vectors served as representative anchors for the metric depth space.
While testing the model, pixel-wise evaluation was conducted in accordance with previous studies. The batch size during training was set to 8.
For the KITTI dataset, the overall data split followed the same protocol as used for NYU Depth V2.
All RGB images, ground-truth depth maps, and Garg masks~\cite{garg2016unsupervised} were resized to $336 \times (336 \times 4)$. Bilinear interpolation was applied to RGB images, while nearest-neighbor interpolation was used for the depth maps and masks. The images were divided into tile-like segments for training. During evaluation, predictions were generated for each tiled input and subsequently merged to form the final output.
In the training phase, the Garg mask was applied to exclude invalid regions. If more than half of the pixels within a patch overlapped with the mask or if the majority of pixels in the patch had missing ground-truth values, that patch was excluded from the loss computation to avoid unreliable supervision.
For evaluation, both the Garg mask and the missing regions in the ground-truth depth maps were excluded from the evaluation, consistent with prior studies. In this case, however, pixel-wise evaluation was conducted rather than patch-wise evaluation to ensure fair comparison.
For the depth table representation, 15 learnable vectors were initialized to correspond to 15 equally spaced depth bins covering the range from 0 to 30 meters, serving as representative anchors for the metric depth space.
The batch size during training was set to 8.

\subsection{Evaluation}
We evaluate the performance of our model using several standard metrics for monocular depth estimation, following the evaluation protocol established by Eigen~\etal~\cite{eigen2014depth}. Let $d_i$ and $d^*_i$ denote the predicted and ground-truth depth at pixel $i$, respectively, and $T$ be the total number of pixels with valid ground truth. The primary error metrics include the Absolute Relative Difference (AbsRel), calculated as $\dfrac{1}{T} \sum_{i} \dfrac{|d_i - d^*_i|}{d^*_i}$, and the Root Mean Square Error (RMSE), defined as $\sqrt{\dfrac{1}{T} \sum_{i} \|d_i - d^*_i\|^2}$. Additionally, we report the Mean Log10 Error ($\log_{10}$), which is computed as $\dfrac{1}{T} \sum_{i} |\log_{10} d_i - \log_{10} d^*_i|$. 
For accuracy assessment, we employ the Threshold Accuracy ($\delta < thr$), which measures the percentage of pixels that satisfy 
$\delta_i = \max \left( \dfrac{d_i}{d^*_i}, \dfrac{d^*_i}{d_i} \right), \quad \text{and} \quad \delta_i < \text{thr}$
for $\text{thr} \in \{1.25, 1.25^2, 1.25^3\}$. While lower values are better for AbsRel, RMSE, and $\log_{10}$, higher values indicate superior performance for the threshold accuracy metrics.

Since we process images in patches, specifically by splitting KITTI images into four horizontal segments, we used horizontal flip-based Test-Time Augmentation (TTA) to utilize the symmetry of the scenes.
TTA was performed during both validation and testing. Specifically, each input image was horizontally flipped, and predictions were obtained for both the original and flipped versions. The two predicted depth maps were then averaged, and evaluation metrics were computed from the resulting averaged depth map.

\subsection{Training}
Regarding the training procedure, the learning rate was initially kept constant without any warm-up.
Specifically, we initialized the learning rates at $3 \times 10^{-4}$ for the Embedding phase and $1 \times 10^{-3}$ for the Depth phase.
This learning rate remained constant until the early-stopping counter reached 20, meaning 20 consecutive epochs without improvement in any validation metric.
Following this, a linear decay was applied, reducing the learning rate to one-tenth of its initial value by the time the counter reached the patience threshold of 50.
Early stopping was not based on the training loss but on the validation metrics computed at every epoch.
This decision was made based on our observation that relying solely on the RMSE loss tended to cause overfitting.
Whenever any of the validation metrics improved, a checkpoint was saved.
Training was terminated when none of the metrics improved for a number of consecutive epochs equal to the patience value, which was set to 50.
Final evaluation was conducted using checkpoints that yielded the optimal results for each respective metric.

\subsection{Network Architecture}

\begin{figure*}
        \centering
        \includegraphics[width=0.97\linewidth]{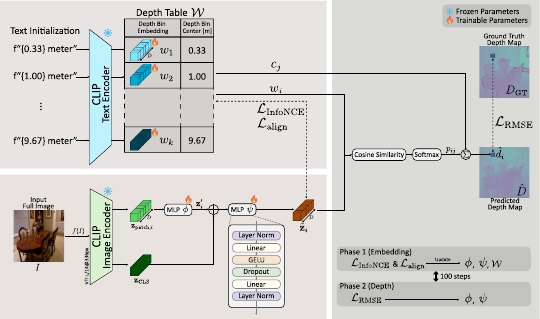}
        \caption{Overview of the proposed architecture. Aside from the CLIP text and image encoders, all operations are performed strictly within the conceptual CLIP embedding space.}
        \label{fig:fig_architecture}
    \end{figure*}

The RGB adapter $\phi$ and the fusion module $\psi$ are both implemented as MLPs consisting of Layer Normalization, two Linear layers with GELU activation, and a Dropout layer.
This specific configuration was selected based on empirical results from preliminary experiments, where it demonstrated the best performance among several architectural candidates.
The detailed layer structure of each module is illustrated in \figref{fig:fig_architecture}.

\subsection{Optimization}
During the training process, the optimization phase was alternated every 100 training steps. 
In the Embedding phase, the model was optimized using the AdamW optimizer
with a learning rate of $3\times10^{-4}$ and a weight decay of $1\times10^{-2}$.
The temperature parameter for the InfoNCE loss was set to $\tau = 0.07$,
and the weighting coefficient was $\lambda_{\infonce} = 1.0$.
In the Depth phase, the AdamW optimizer was also used
with a learning rate of $1\times10^{-3}$ and no weight decay to minimize the RMSE loss.

All experiments were conducted on the TSUBAME 4.0 supercomputer at the Institute of Science Tokyo.
A quarter node was used for training and evaluation. The node was equipped with an AMD EPYC 9654 CPU and a single NVIDIA H100 SXM5 GPU.
Under this configuration, the training and evaluation process for each model typically required approximately 24 to 48 hours to complete.
\section{Results}
\subsection{Quantitative Results}

\begin{table*}[!htb]
  \centering
  \caption{Depth estimation results on NYU Depth V2 (Eigen crop).}
  \label{tab:nyu_results}
  \resizebox{\textwidth}{!}{%
    \begin{tabular}{lrrrrrr}
      \toprule
      Model                                    & AbsRel $\downarrow$ & RMSE $\downarrow$ & $\log_{10}$ $\downarrow$ & $\delta<1.25$ $\uparrow$ & $\delta<1.25^2$ $\uparrow$ & $\delta<1.25^3$ $\uparrow$ \\
      \midrule
      DepthCLIP (Zhang~\etal)~\cite{zhang2022can}   & 0.388  & 1.167 & 0.156 & 0.394 & 0.683 & 0.851 \\
      Hu~\etal~\cite{hu2024learning} & 0.347  & 1.049 & 0.140 & 0.428 & 0.732 & 0.898 \\
      Auty \& Mikolajczyk~\cite{auty2023learning}              & 0.319  & 0.970 & 0.128 & 0.465 & 0.776 & 0.922 \\
      \midrule
      \rowcolor{gray!20}
      \textbf{PureCLIP-Depth (Ours)}                     & \textbf{0.201} & \textbf{0.670} & \textbf{0.084} & \textbf{0.671} & \textbf{0.917} & \textbf{0.979} \\
      \bottomrule
    \end{tabular}%
  }
\end{table*}

\begin{table*}[!htb]
  \centering
  \caption{Depth estimation results on KITTI (Garg mask).}
  \label{tab:kitti_results}
  \resizebox{\textwidth}{!}{%
    \begin{tabular}{lrrrrrr}
      \toprule
      Model & AbsRel $\downarrow$ & RMSE $\downarrow$ & $\log_{10}$ $\downarrow$ & $\delta<1.25$ $\uparrow$ & $\delta<1.25^2$ $\uparrow$ & $\delta<1.25^3$ $\uparrow$ \\
      \midrule
      DepthCLIP (Zhang~\etal)~\cite{zhang2022can} & 0.473 & 12.958 & 0.680 & 0.281 & 0.531 & 0.696 \\
      Hu~\etal~\cite{hu2024learning}              & 0.384 & 12.290 & 0.632 & 0.312 & 0.569 & 0.739 \\
      Auty \& Mikolajczyk~\cite{auty2023learning}              & 0.238 & 5.756  & 0.088 & 0.652 & 0.877 & 0.957 \\
      \midrule
      \rowcolor{gray!20}
      \textbf{PureCLIP-Depth (Ours)} & \textbf{0.172} & \textbf{1.062} & \textbf{0.073} & \textbf{0.739} & \textbf{0.949} & \textbf{0.988} \\
      \bottomrule
    \end{tabular}%
  }
\end{table*}
The proposed method outperformed all previous approaches~\cite{zhang2022can,hu2024learning,auty2023learning} across all evaluation metrics as shown in \tabref{tab:nyu_results}.
On the NYU Depth V2 dataset, the improvement in $\delta<1.25$ (from 0.465 to 0.671, $+44\%$) indicates a significant enhancement in fine-grained structural recognition and local depth precision.
The consistent reduction in AbsRel, RMSE, and $\log_{10}$ errors further demonstrates that the model achieves more stable and accurate depth estimation across different distance ranges.
Similarly, on the KITTI dataset, a comparable trend was observed, where the proposed method maintained high accuracy even in large-scale outdoor scenes with long-range depth variations.
These results demonstrate that the proposed approach possesses strong representational capability and robustness across different environments ranging from indoor scenes up to 10 meters to outdoor scenes exceeding 30 meters.

    \begin{figure*}[!h]
        \centering
        \includegraphics[width=1.0\linewidth]{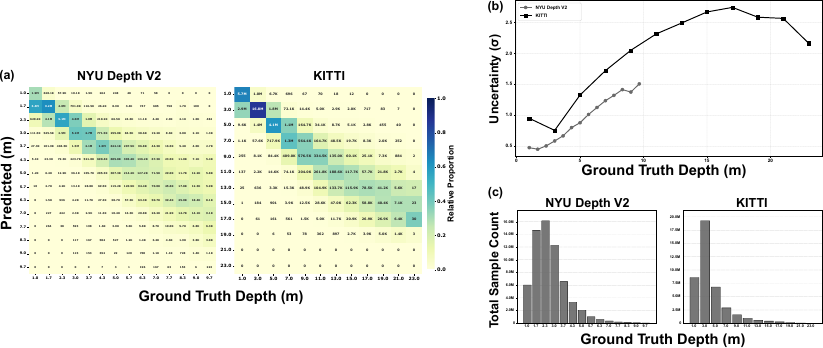}
        \caption{Analysis of depth estimation performance across different distance ranges. (a) Joint distributions of predicted versus ground-truth depth for the NYU Depth V2 and KITTI datasets. (b) Empirical uncertainty, calculated as the standard deviation ($\sigma$), with respect to the ground-truth depth. (c) Histograms illustrating the distribution of ground-truth depth (total sample count) in each dataset. All plots are generated within the ranges of 1.0 to 9.7~m for NYU Depth V2 and 1.0 to 23.0~m for KITTI; the initial bin for NYU (0.33~m) and far-range bins for KITTI (25.0, 27.0, and 29.0~m) are trimmed due to the absence of sufficient ground-truth samples in those intervals.}
        \label{fig:fig_heatmap}
    \end{figure*}

To further investigate the relationship between ground-truth values and the model's predictions, we analyzed the joint distribution, empirical uncertainty, and dataset density as shown in \figref{fig:fig_heatmap}. 
As shown in \figref{fig:fig_heatmap}(a), the predictions for both NYU Depth V2 and KITTI exhibit a high concentration along the diagonal across a wide range of distances. This indicates that the proposed mapping within the CLIP embedding space maintains high inference accuracy and a strong consistency between the estimated and true depth values, even without a traditional decoder architecture.

Conversely, \figref{fig:fig_heatmap}(b) reveals that the variance of the inference results increases significantly at larger depths. While this could be interpreted as an inherent limitation of the model's capacity at far ranges, a comparison with the ground-truth distribution in \figref{fig:fig_heatmap}(c) suggests a more data-driven cause. The histograms show a drastic reduction in the number of samples in far-range regions, indicating that the lack of sufficient supervision signals during training leads to the observed instability and increased uncertainty in the model’s estimations.

Furthermore, in the far-range regions of \figref{fig:fig_heatmap}(a), a systematic negative bias is observed where the model tends to underestimate the depth. This phenomenon can be attributed to the increased uncertainty and the imbalanced sample density of the datasets. When the model faces high uncertainty due to sparse far-range data, the optimization process intended to minimize the expected loss biases the model toward high-density regions, a statistical effect known as "regression to the mean." Consequently, the model tends to predict lower depth values at greater distances.

\subsection{Qualitative Results}
    \begin{figure*}[!h]
        \centering
        \includegraphics[width=1.0\linewidth]{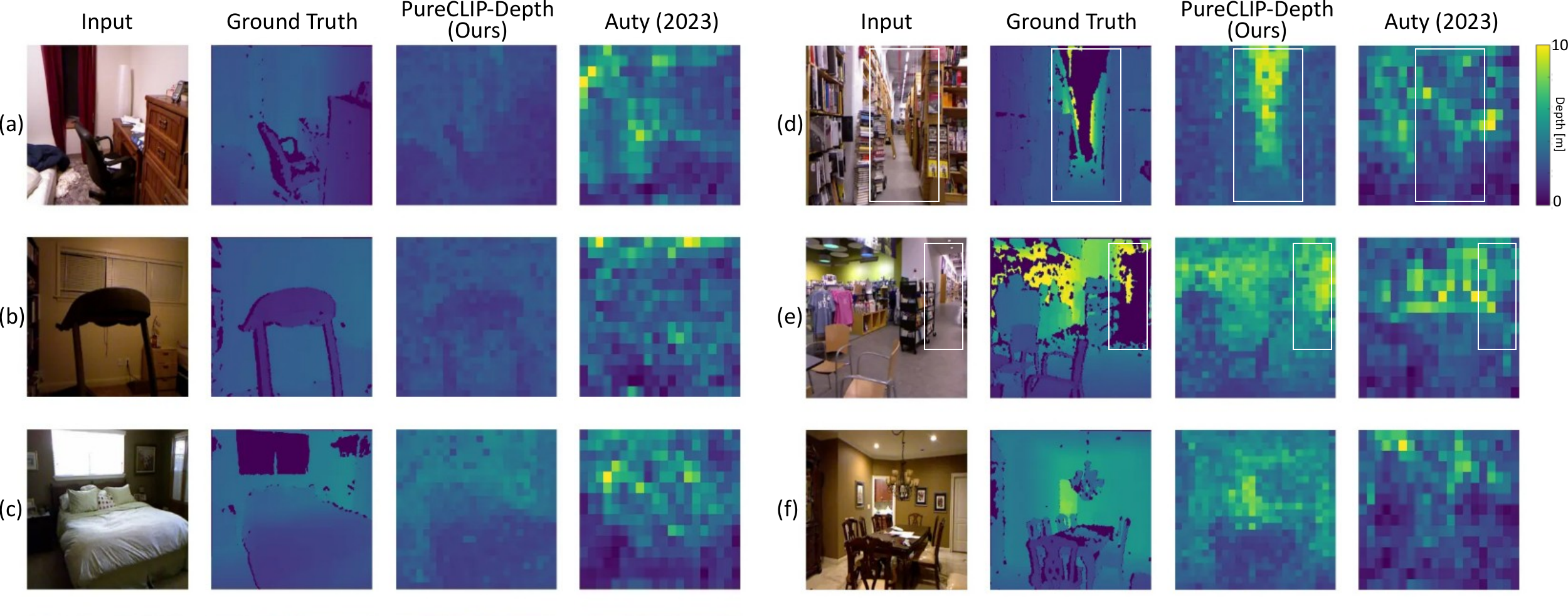}
        \caption{Qualitative visualization on NYU Depth V2 test set. Compared to Auty \& Mikolajczyk~\cite{auty2023learning}, our method shows significant improvements in depth quality for (a, b) chair shapes, (c) depth variations across the bed, (d) regions with missing values in the library aisle, (e) missing values in the shop background, and (f) the central desk and the entrance to the back room.}
        \label{fig:fig_quantitative_nyu}
    \end{figure*}
\figref{fig:fig_quantitative_nyu} illustrates the qualitative visualizations on the NYU Depth V2 dataset. 
It is evident that the proposed method improves the preservation of object boundaries and fine-grained structural details compared to the approach of Auty \& Mikolajczyk~\cite{auty2023learning}. For example, the shapes of tables in \figref{fig:fig_quantitative_nyu}(b) and the outlines of chairs in \figref{fig:fig_quantitative_nyu}(e) are reconstructed more clearly. In the case of chairs, depth differences between components such as armrests and the background are maintained, resulting in object-consistent depth structures. 

In addition to these structural improvements,
the proposed method produces depth predictions whose range varies across scenes. 
In cases with significant depth variations, such as \figref{fig:fig_quantitative_nyu}(d), the predicted values span a wider range, whereas in \figref{fig:fig_quantitative_nyu}(b), they are concentrated within a narrower range.
These results indicate that the proposed method does not output a uniform depth distribution, but instead exhibits depth ranges that adapt to the underlying scene structure.

Furthermore, even in regions where training data are frequently missing due to sensor limitations, such as the central aisle in \figref{fig:fig_quantitative_nyu}(d) or the distant background in \figref{fig:fig_quantitative_nyu}(e)—the proposed method produces depth predictions that remain consistent with the surrounding context.
Although these regions are excluded from quantitative evaluation, the visualizations suggest that the model produces globally coherent depth estimates aligned with the overall scene structure.

    \begin{figure*}[!h]
        \centering
        \includegraphics[width=1.0\linewidth]{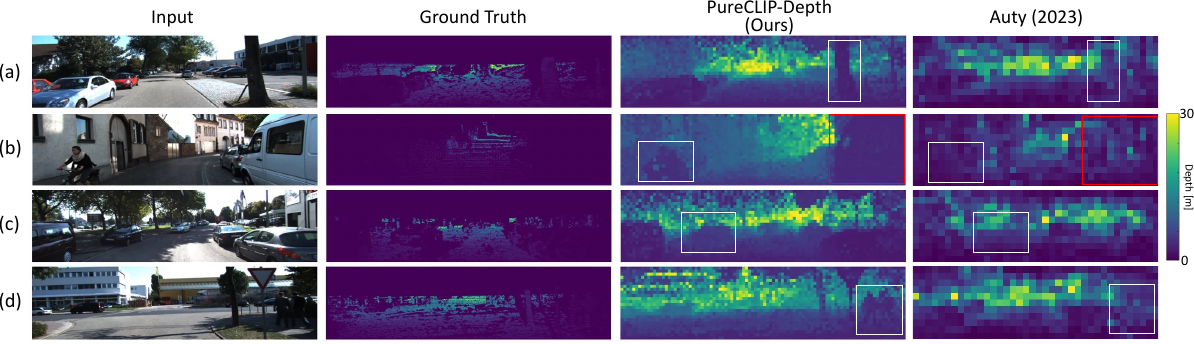}
        \caption{Qualitative visualization on KITTI test set. Compared to Auty \& Mikolajczyk~\cite{auty2023learning}, our method more accurately reflects depth for (a) the shapes of street trees, (b) cyclists, (c) parked cars, and (d) pedestrians in the shade.}
        \label{fig:fig_quantitative_kitti}
    \end{figure*}
\figref{fig:fig_quantitative_kitti} presents qualitative comparisons on the KITTI dataset. Compared to the results of Auty \& Mikolajczyk~\cite{auty2023learning}, the proposed method exhibits finer structural fidelity, particularly for thin objects such as trees, utility poles, and traffic signs.
For vehicles, spatially coherent depth predictions are obtained not only when the entire object is visible in the image but also when only partial views are available, as indicated by the red boxes in \figref{fig:fig_quantitative_kitti}(b). This behavior shows that the model makes use of surrounding context when inferring depth. 
In addition, relatively small objects such as the pedestrians in \figref{fig:fig_quantitative_kitti}(d) and the cyclists in \figref{fig:fig_quantitative_kitti}(b) are predicted with depth values that are clearly separated from the background.
Even in challenging cases where pedestrians wearing dark clothing appear in shadowed regions, the target regions are distinguished from their surroundings. These results indicate that the model maintains structural integrity even under unfavorable illumination conditions.

Despite these qualitative successes, consistent depth estimation for sky regions remains a common challenge due to the limitations of the dataset. Theoretically, sky regions correspond to infinite depth.
However, LiDAR-based ground-truth depth annotations are not available for all frames in these regions. As a result, sky regions are inherently difficult areas for consistent depth estimation. Since such regions are commonly excluded from evaluation using the Garg mask and are of limited importance in control and navigation applications, we restrict our discussion to qualitative observations regarding the difficulty of depth prediction in these regions.

\subsection{Ablation Study}

\begin{table*}[!htb]
  \centering
  \caption{Ablation study on Rotation MLPs ($\phi, \psi$).}
  \label{tab:ablation_mlp}
  \resizebox{\textwidth}{!}{%
    \begin{tabular}{llrrrrrr}
      \toprule
      Dataset & Setting
      & AbsRel $\downarrow$ & RMSE $\downarrow$ & $\log_{10}$ $\downarrow$
      & $\delta<1.25$ $\uparrow$ & $\delta<1.25^2$ $\uparrow$ & $\delta<1.25^3$ $\uparrow$ \\
      \midrule
      NYU   & w/o MLPs  & 0.320 & 1.015 & 0.134 & 0.443 & 0.751 & 0.909 \\
      \rowcolor{gray!20}
      NYU   & with MLPs (Baseline)
            & \textbf{0.201} & \textbf{0.670} & \textbf{0.084}
            & \textbf{0.671} & \textbf{0.917} & \textbf{0.979} \\
      \midrule
      KITTI & w/o MLPs & 0.371 & 1.889 & 0.147 & 0.405 & 0.727 & 0.886 \\
      \rowcolor{gray!20}
      KITTI & with MLPs (Baseline)
            & \textbf{0.172} & \textbf{1.062} & \textbf{0.073}
            & \textbf{0.739} & \textbf{0.949} & \textbf{0.988} \\
      \bottomrule
    \end{tabular}%
  }
\end{table*}

To analyze the effect of the rotation MLPs ($\phi, \psi$), we conducted an ablation study in which all MLP parameters in the proposed model were frozen at their initial random values. Specifically, gradient computation was disabled for the MLP modules, and their parameters were excluded from the optimizer updates.
As shown in \tabref{tab:ablation_mlp}, removing these MLPs leads to a consistent performance drop across all metrics on both NYU Depth V2 and KITTI. The MLPs are responsible for rotating RGB embeddings toward the depth direction within the CLIP embedding space, and freezing them removes the model’s ability to adapt the embeddings. As a result, the model relies only on the fixed CLIP representations together with the rearrangement of the learnable depth table, which limits its expressive capacity.
Nevertheless, the model without the MLPs still achieves reasonable performance. This is because the depth table vectors $\mathcal{W}$ remain learnable, allowing the model to partially align fixed input embeddings by adapting the reference depth vectors. Similar behavior has been observed in prior work by Hu~\etal~\cite{hu2024learning}. In contrast, the clear performance gap quantified in this ablation highlights the effectiveness of the proposed rotation-based alignment introduced by the rotation MLPs.

\begin{table*}[!htb]
  \centering
  \caption{Ablation study on the global context token $\mathbf{z}_{\text{CLS}}$.}
  \label{tab:ablation_cls}
  \resizebox{\textwidth}{!}{%
    \begin{tabular}{llrrrrrr}
      \toprule
      Dataset & Setting
      & AbsRel $\downarrow$ & RMSE $\downarrow$ & $\log_{10}$ $\downarrow$
      & $\delta<1.25$ $\uparrow$ & $\delta<1.25^2$ $\uparrow$ & $\delta<1.25^3$ $\uparrow$ \\
      \midrule
      NYU   & w/o CLS & 0.215 & 0.698 & 0.088 & 0.655 & 0.907 & 0.975 \\
      \rowcolor{gray!20}
      NYU   & with CLS (Baseline)
            & \textbf{0.201} & \textbf{0.670} & \textbf{0.084}
            & \textbf{0.671} & \textbf{0.917} & \textbf{0.979} \\
      \midrule
      KITTI & w/o CLS & 0.182 & 1.111 & 0.076 & 0.717 & 0.940 & 0.986 \\
      \rowcolor{gray!20}
      KITTI & with CLS (Baseline)
            & \textbf{0.172} & \textbf{1.062} & \textbf{0.073}
            & \textbf{0.739} & \textbf{0.949} & \textbf{0.988} \\
      \bottomrule
    \end{tabular}%
  }
\end{table*}
Subsequently, to evaluate the effect of incorporating global context, we investigated the impact of the CLS token. While the proposed method utilizes both patch-level embeddings and the CLS embedding from CLIP to capture global image information, this ablation removes the concatenation of the CLS representation while keeping the rest of the architecture unchanged. To account for the absence of concatenation, the input dimension of the first linear layer in the second MLP was adjusted from $2D$ to $D$.
As shown in \tabref{tab:ablation_cls}, the inclusion of the CLS token leads to small yet consistent improvements across all evaluation metrics on both datasets. The fact that all metrics are improved suggests that the CLS representation provides a stable positive contribution to depth estimation. In particular, the $\delta<1.25$ metric improves by approximately 0.02 on both NYU Depth V2 and KITTI, indicating that the global CLS embedding and local patch embeddings are meaningfully fused through the MLP, thereby enhancing the accuracy of depth prediction.

\begin{table*}[!htb]
  \centering
\caption{Ablation study on loss functions: $\mathcal{L}_{\infonce}$, $\mathcal{L}_{\myalign}$, and $\mathcal{L}_{\rmse}$.}
\label{tab:ablation_loss}
  \resizebox{\textwidth}{!}{%
    \begin{tabular}{llrrrrrr}
      \toprule
      Dataset & Loss
      & AbsRel $\downarrow$ & RMSE $\downarrow$ & $\log_{10}$ $\downarrow$
      & $\delta<1.25$ $\uparrow$ & $\delta<1.25^2$ $\uparrow$ & $\delta<1.25^3$ $\uparrow$ \\
      \midrule
      NYU   & InfoNCE
            & 0.216 & 0.693 & 0.087 & 0.658 & 0.908 & 0.974 \\
      NYU   & InfoNCE + alignment
            & 0.210 & 0.687 & 0.087 & 0.654 & 0.909 & 0.976 \\
      \rowcolor{gray!20}
      NYU   & InfoNCE + alignment + RMSE (Baseline)
            & \textbf{0.201} & \textbf{0.670} & \textbf{0.084}
            & \textbf{0.671} & \textbf{0.917} & \textbf{0.979} \\
      \midrule
      KITTI & InfoNCE
            & 0.215 & 1.145 & 0.099 & 0.585 & 0.867 & 0.980 \\
      KITTI & InfoNCE + alignment
            & 0.216 & 1.138 & 0.097 & 0.588 & 0.882 & 0.980 \\
      \rowcolor{gray!20}
      KITTI & InfoNCE + alignment + RMSE (Baseline)
            & \textbf{0.172} & \textbf{1.062} & \textbf{0.073}
            & \textbf{0.739} & \textbf{0.949} & \textbf{0.988} \\
      \bottomrule
    \end{tabular}%
  }
\end{table*}
Finally, to examine the influence of each loss component, we assessed the model's performance under various loss settings. 
For the two ablated settings without the RMSE loss, the alternating optimization strategy was not applied, and the models were trained using a single loss formulation.
As shown in \tabref{tab:ablation_loss}, the difference between using InfoNCE alone and InfoNCE + alignment loss is relatively small. This can be attributed to the fact that the contrastive objective of InfoNCE inherently encourages embeddings to move closer to the target depth table vector while separating them from others, resulting in a partially overlapping effect with the alignment loss. Nevertheless, since the additional computational cost of the alignment loss is negligible and small improvements in RMSE and $\delta<1.25$ are observed, we adopt both losses in the proposed method.

In contrast, incorporating the RMSE loss with the alternating optimization strategy leads to consistent and clear improvements across all evaluation metrics, with particularly notable gains on the KITTI dataset. While InfoNCE and alignment losses primarily focus on optimizing the embedding representations, the RMSE loss directly penalizes errors in the predicted depth maps. As a result, it functions as an independent and complementary objective that has a stronger impact on the final depth estimation performance.

\section{Conclusion}
In this paper, we proposed PureCLIP-Depth, a purely conceptual, encoder-only approach for MDE that maps RGB directly to depth within the CLIP embedding space, completely eliminating the reliance on explicit text prompts.
By introducing a learnable depth table and an alternating optimization strategy, our simple architecture achieved state-of-the-art performance among CLIP encoder-only models across all quantitative evaluation metrics on both indoor and outdoor datasets.
Notably, we empirically demonstrated the existence of a learnable mapping function from RGB to depth within the CLIP embedding space. 
Qualitative results further supported these quantitative achievements, demonstrating that our method effectively preserves fine-grained structural details and maintains global depth consistency, even in regions prone to missing values.
Furthermore, our ablation studies numerically validated the core components of our method, proving the critical effectiveness of the MLP-based spatial rotation, the integration of the CLS token for global context, and the proposed loss function strategy.

Despite these promising results, addressing the prediction uncertainty and underestimation at long ranges caused by dataset imbalance remains a challenge for future work. Additionally, leveraging our method's unique independence from geometric features, we aim to explore its applicability to low-contrast images with blurred outlines, such as thermal infrared images.

\section*{Acknowledgments}
This work was supported by JST SPRING, Japan Grant Number JPMJSP2106, and was carried out using the TSUBAME4.0 supercomputer at Institute of Science Tokyo.

\printbibliography

\end{document}